\documentclass[sigconf]{acmart}
\AtBeginDocument{%
  }

\setcopyright{acmlicensed}
\copyrightyear{2018}
\acmYear{2018}
\acmDOI{XXXXXXX.XXXXXXX}
\acmConference[Conference acronym 'XX]{Make sure to enter the correct
  conference title from your rights confirmation email}{June 03--05,
  2018}{Woodstock, NY}
\acmISBN{978-1-4503-XXXX-X/2018/06}

\usepackage{adjustbox}
\usepackage{multirow}
\begin{document}

\title[Thou Shalt Not Prompt: Zero-Shot HAR in Smart Homes via Language Modeling of Sensor Data \& Activities]{
Thou Shalt Not Prompt: Zero-Shot Human Activity Recognition in Smart Homes via Language Modeling of Sensor Data \& Activities
}

\author{Sourish Gunesh Dhekane}
\email{sourish.dhekane@gatech.edu}
\orcid{https://orcid.org/0009-0004-0035-819X}
\affiliation{%
  \institution{School of Interactive Computing, Georgia Institute of Technology}
  \city{Atlanta}
  \country{USA}
}

\author{Thomas Ploetz}
\email{thomas.ploetz@gatech.edu}
\orcid{https://orcid.org/0000-0002-1243-7563}
\affiliation{%
  \institution{School of Interactive Computing, Georgia Institute of Technology}
  \city{Atlanta}
  \country{USA}}

\renewcommand{\shortauthors}{Trovato et al.}

\begin{abstract}
Developing zero-shot human activity recognition (HAR) methods is a critical direction in smart home research -- considering its impact on making HAR systems work across smart homes having diverse sensing modalities, layouts, and activities of interest.
The state-of-the-art solutions along this direction are based on generating natural language descriptions of the sensor data and feeding it via a carefully crafted prompt to the LLM to perform classification.    
Despite their performance guarantees, such ``prompt-the-LLM'' approaches carry several risks, including privacy invasion, reliance on an external service, and inconsistent predictions due to version changes, making a case for alternative zero-shot HAR methods that do not require prompting the LLMs. 
In this paper, we propose one such solution that models sensor data and activities using natural language, leveraging its embeddings to perform zero-shot classification and thereby bypassing the need to prompt the LLMs for activity predictions. 
The impact of our work lies in presenting a detailed case study on six datasets, highlighting how language modeling can bolster HAR systems in zero-shot recognition.    

\end{abstract}

\ccsdesc[500]{Human-centered computing~Ubiquitous and mobile computing}
\ccsdesc[300]{Computing methodologies~Machine learning}
\ccsdesc[500]{Human-centered computing~Empirical studies in ubiquitous and mobile computing}

\keywords{human activity recognition, zero-shot classification, smart homes}



\begin{teaserfigure}
  \includegraphics[width=\textwidth]{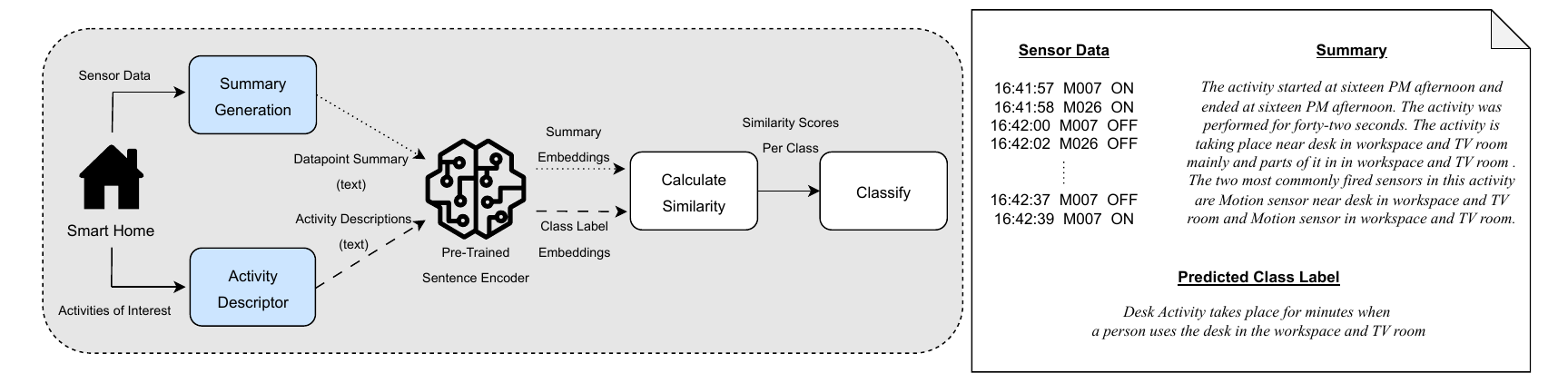}
  \caption{The proposed zero-shot solution models sensor data and activities of interest using natural language via the novel ``Summary Generation'' and ``Activity Descriptor'' modules (colored blue). 
  For zero-shot classification, the similarity between the embeddings of the resultant summary and activity descriptions is compared --  the label with the highest similarity is predicted during inference. 
  An example of the sensor datapoint, its summary, and the predicted class label is shown on the right.  
  }
  \label{fig:teaser}
\end{teaserfigure}

\received{20 February 2007}
\received[revised]{12 March 2009}
\received[accepted]{5 June 2009}

\maketitle

\section{Introduction}
Building a successful human activity recognition (HAR) system in smart homes is synonymous with having the capacities of: (\textit{i}) \textbf{Generalizable Recognition}: the ability to model heterogeneous multimodal sensor data across smart homes with different layouts, sensor setups, activities of interest, and user behaviors, and (\textit{ii}) \textbf{Zero-Shot Recognition}: the ability to recognize activities without collecting and annotating large data amounts
-- a goal that is long sought by the smart home HAR community to cater to a growing home automation industry.

Towards achieving this goal, the approach of generating a textual description of sensor data and then prompting it to the Large Language Models (LLMs) for classification has proven to be an effective state-of-the-art (SOTA) solution \cite{civitarese2024large, gao2024unsupervised, xia2023unsupervised, takeda2024synergistic}. 
By converting sensor events into text descriptions, data from any ambient sensing modality and across any sensor setup can be incorporated into HAR systems, improving the generalizability across sensing infrastructures. 
Further, commonsense and world knowledge from LLMs is leveraged for classification -- leading to the zero-shot capability for predicting any set of activities. 
However, these ``prompt-the-LLM'' based approaches have several shortcomings. 

First, these approaches rely entirely on predictions made by an LLM, a service outside of the smart home ecosystem. 
Thus, in situations where the LLM service can be denied (network issues, service outages, etc.), there exists a danger of the entire HAR system coming to a standstill, resulting in undesirable outcomes for upstream HAR-based applications for sensitive use cases, especially in healthcare.
Also, considering the privacy requirements of such sensitive use cases, many users may not be willing to share their in-home data with an outside party. 
Finally, the LLMs are notorious when it comes to their unpredictability in the generated outcomes -- a study conducted by Takeda et al. \cite{takeda2024synergistic} observed that the same LLM prompt queried across different (upgrading) LLM versions produced degrading HAR performance.
This inspires us to explore language modeling of sensor data to avoid said pitfalls while keeping the HAR system generalizable and zero-shot in nature.

Language modeling refers to generating textual embeddings of sensor data (or its descriptions) and using them as the representation space to perform HAR \cite{thukral2025layout}. 
In this work, we propose a
generalizable and zero-shot HAR solution in smart homes that leverages language modeling of sensor data and activity labels to bypass the need to rely on LLMs.
Our contribution lies in performing language modeling at the correct granularity by converting sensor data to its text summary and activity labels to their text descriptions. 
The embeddings of descriptions are then compared against that of the summary using a similarity metric, where the activity label corresponding to the embedding with the highest similarity is inferred during zero-shot classification. 
The workflow of this solution is shown in Figure \ref{fig:teaser}. 

To evaluate our approach, we cover six datasets (Aruba, Milan, Cairo, Kyoto7, MARBLE, and ARAS) with diverse sensing modalities, layouts, and activities of interest, which showcase the generalizability of our approach.
Our solution, consistent in its structure across all datasets, achieves comparable performance to three LLM-based SOTA baselines with their underlying prompts tuned specifically to the choice of their datasets.   
We also evaluate our method in a few-shot scenario, where we achieve significant performance improvements in five out of six datasets in our analysis, which highlights the compatibility of our solution to be used in real-world human-in-the-loop HAR systems.  
Finally, we perform an ablation study to justify the design choices made in our solution, thereby guiding the optimal design of our HAR system.

The impact of our work lies in presenting a detailed case study of how language modeling can bolster the HAR systems in tackling the challenges of generalizable and zero-shot recognition. 
Our discussion on how to use language modeling correctly, while keeping in mind its potential limitations, would provide the smart home HAR community with necessary guidelines in their future explorations for use cases in monitoring, healthcare, and assisted living.    

\section{Related Works}
Developing HAR systems for smart homes using minimum data and annotations is an active research direction, where the SOTA draws inspiration from the works in transfer learning \cite{dhekane2025transfer}.  
While methods like self-supervision \cite{chen2024enhancing} or adversarial training \cite{sanabria2021unsupervised} help reduce the annotation need, they still require large amounts of unlabeled data, which may not be feasible in newer smart home settings, where HAR systems are required to be deployed on the spot. 
This gives rise to the requirements of generalizability and zero-shot recognition capabilities in HAR systems for smart homes. 

Language modeling has proven to be an effective solution to satisfy both these requirements. 
Methods like Word2Vec \cite{yu2023fine, niu2022source} and Neural word embedding \cite{azkune2020cross} have been used to convert raw sensor data into their respective embeddings. 
However, to learn a mapping between these embeddings and activity labels, supervised training is required on every distinct smart home setting.
This requirement can be removed by generating textual descriptions of sensor triggers and using their resultant embeddings to train a generic classifier that can be transferred to different smart home settings \cite{thukral2025layout}. 
However, these works do not recognize activities in a zero-shot manner. 

With the advent of LLMs, it has become possible to have both generalizability as well as zero-shot recognition capabilities in HAR systems. 
In particular, different methods have been developed that prompted LLMs in a specific manner, such that their commonsense and world knowledge were leveraged to recognize activities without any additional training \cite{xia2023unsupervised, fritsch2024hierarchical, takeda2024synergistic}.
These methods can be easily extended to few-shot settings, where labeled data samples can be fed to the LLM for finetuning \cite{cleland2024leveraging}. 
Also, these methods have been modified by either pre-processing the sensor data into a more enhanced format for the LLM to correctly predict activity labels \cite{civitarese2024large}, or by using LLM predictions to perform iterative training of a classifier \cite{gao2024unsupervised}.
A common thread in all these works is that they rely entirely on LLM inference, which can be privacy invasive and proven to be unreliable in its predictions \cite{takeda2024synergistic}. 

To address this shortcoming, we propose removing LLM inference from the HAR pipeline and replacing it with embedding space arithmetic between sensor data summaries and activity descriptions. 
To the best of our knowledge, only the work by Fritsch et al. \cite{fritsch2024hierarchical} explores this possibility; however, they do not include activity descriptions in their method, whereas our work leverages both sensor data summaries and activity descriptions to provide a comprehensive case study with SOTA comparisons and ablations.

\section{Methodology}
\label{sec:method}
Our proposed zero-shot solution is based on the principle of language modeling of sensor data $X = \{X_1, X_2,...,X_n\}$ and activity labels $Y = \{Y_1, Y_2,...,Y_k\}$, which can be viewed as a four-step process:
(\textit{i}) Generating textual descriptions of the sensor data $T^X = \{T^X_1, T^X_2,...,T^X_n\}$ and activity labels $T^Y = \{T^Y_1, T^Y_2,...,T^Y_k\}$ at the correct granularity, (\textit{ii}) Generating embeddings of these textual descriptions $E^X = \{E^X_1, E^X_2,...,E^X_n\}$ and $E^Y = \{E^Y_1, E^Y_2,...,E^Y_k\}$ using pre-trained sentence encoder (\textit{iii}) Calculating similarity between $E^X_i$ and $\{E^Y\}$ for each datapoint $X_i$, and (\textit{iv}) Assigning class label $Y_j$ to datapoint $X_i$ based on highest similarity. 
In this process, we use only a publicly available pre-trained sentence encoder, and no labeled or unlabeled sensor data is required for any additional (pre)training of the HAR system, making it a zero-shot solution. 
Our novel contributions lie in generating said $T^X$ and $T^Y$ at the correct granularity.


\subsection{Summary Generation $(T^X)$}
\label{subsec:summary_generation}
The goal of the summary generation component is to construct a textual summary $T^X_i$ of the sensor datapoint $X_i$ such that it captures the essence of the activity being performed.
To design a single consistent \textit{skeleton} of our summary, we take into consideration the information present in the floorplans, sensor layouts, and activities of interest for all datasets in our study.
This skeleton consists of four types of information. 

\subsubsection{Time of Occurrence}
We observe that many activities of interest like \{Morning\_Meds, Evening\_Meds\}, \{Breakfast, Lunch, Dinner, Night\_Wandering\}, and \{Preparing/Having\_Breakfast, Preparing/Having\_Lunch, Preparing/Having\_Dinner\}, are heavily dependent on the time at which they were performed. 
Hence, we include a sentence in the summary having the following structure:
    \begin{quote}
        \texttt{The activity started at <24-Hr Time> <AM/PM> <Period of Day> and ended at <24-Hr Time> <AM/PM> <Period of Day>}
    \end{quote}

\subsubsection{Duration of Activity}
The duration of activity occurrence plays a key role in identifying many activities, such as Sleeping and Relaxing tend to take longer (over minutes or hours), whereas activities like Bed\_to\_Toilet and Entering/Leaving\_Home occur over much shorter durations (over seconds). 
Hence, we include a sentence with the following structure in our summary:
    \begin{quote}
        \texttt{The activity was performed for <Duration> <Unit>}
    \end{quote} 

\subsubsection{Top-\textit{k} Locations of Occurrence}
Many smart home datasets contain data originating from event-based sensors, which encode information on movement patterns of the resident from one location to another -- extraction of which can significantly boost the zero-shot recognition performance. 
Hence, we add a sentence with the following structure in our summaries (value of $k$ set at two, by default):
    \begin{quote}
        \texttt{The activity is taking place <Most Common Location> mainly and parts of it in <Second Most Common Location>}
    \end{quote} 
The location of each sensor is retrieved by referring to the layout of the smart homes -- by setting location tags 
such as, in the kitchen near the stove, in the bedroom near the bed, etc. 

\subsubsection{Top-\textit{k} Sensors}
Similarly to using top-\textit{k} locations, the top-\textit{k} sensors fired in a sensor window can also be used to improve summary generation. 
The key idea here is to not only provide which sensing modalities (e.g., motion, door, magnetic, smart plug, etc.) were fired the most, but also their context, i.e., location or attached-to-object information, such as magnetic sensor on the medicine cabinet door.  
Thus, we add a sentence having the following structure in our summaries:
    \begin{quote}
        \texttt{The two most commonly fired sensors in this activity are <Sensor 1> <Context of Sensor 1> and <Sensor 2> <Context of Sensor 2>}
    \end{quote} 

\subsubsection{Special Cases}
\label{subsubsec:special_cases}
In the zero-shot recognition of certain \textit{special} activities, the summaries generated by our approach may not correctly reflect the activity-specific signatures. 
This is when we add special rules in our summary generation module using common-sense heuristic knowledge about the nature of those special activities. 
For example, in the MARBLE dataset, the activities of Answering\_Phone and Making\_Phone\_Call are characterized by the resident using their smartphone (pressing the make-call or answer-call buttons). 
However, their summaries are extremely unlikely to contain this specific information, and hence, 
we make special rules that are grounded in the common-sense knowledge of certain activities. 
For example, if the sensors on smartphones are observed in the window, we mention them in our Top-\textit{k} sensors sentence by default. 
These special rules are derived prior to the zero-shot recognition process, without any labeled datapoints as reference. 


\subsection{Activity Description $(T^Y)$}
\label{subsec:activity_description}
To perform zero-shot classification, we focus on generating descriptions $T^Y$ that \textit{precisely} describe the underlying activity in the context of the smart home. 
The goal of this step is to generate activity descriptions that are more likely to be close to data point summaries $T^X$ in the language embedding space. 
To generate such descriptions, we leverage the smart home layout as well as the metadata provided in the datasets to understand what each activity denotes precisely. 
For example, the sensor layout of the Milan dataset shows a desk situated in the workspace and the TV room, which helps us describe the Desk\_Activity as:  
\begin{quote}
\texttt{Desk Activity takes place for minutes when a person uses the desk in the workspace and TV room}        
\end{quote} 
In case of the MARBLE dataset, the metadata contains activity descriptions, e.g., \textit{Watching TV: "the actor switches on the television if is off and sits on a living room chair; at the end of the activity, he/she can switch off the television if nobody else is watching it"}. 
Using this metadata, we arrive at the following description:
\begin{quote}
\texttt{Watching TV activity takes place in the living room for minutes or hours when a person turns on the television TV and sits on a living room couch
}    
\end{quote}

In this work, we limit the activity descriptions to one sentence each. 
In that sentence, we mention key details like: (\textit{i}) how long the activity is likely to be performed, (\textit{ii}) what location it is likely to be performed, and (\textit{iii}) any special sensor reading occurrences that act as a signature to that activity (e.g., the smartphone sensors for the Answering\_Phone and Making\_Phone\_Call activities in the MARBLE dataset), in that order. 
These descriptions are derived solely from observing the smart home layouts and the metadata available with the datasets. 


\begin{figure*}[t]
  \centering
  \includegraphics[width=\textwidth]{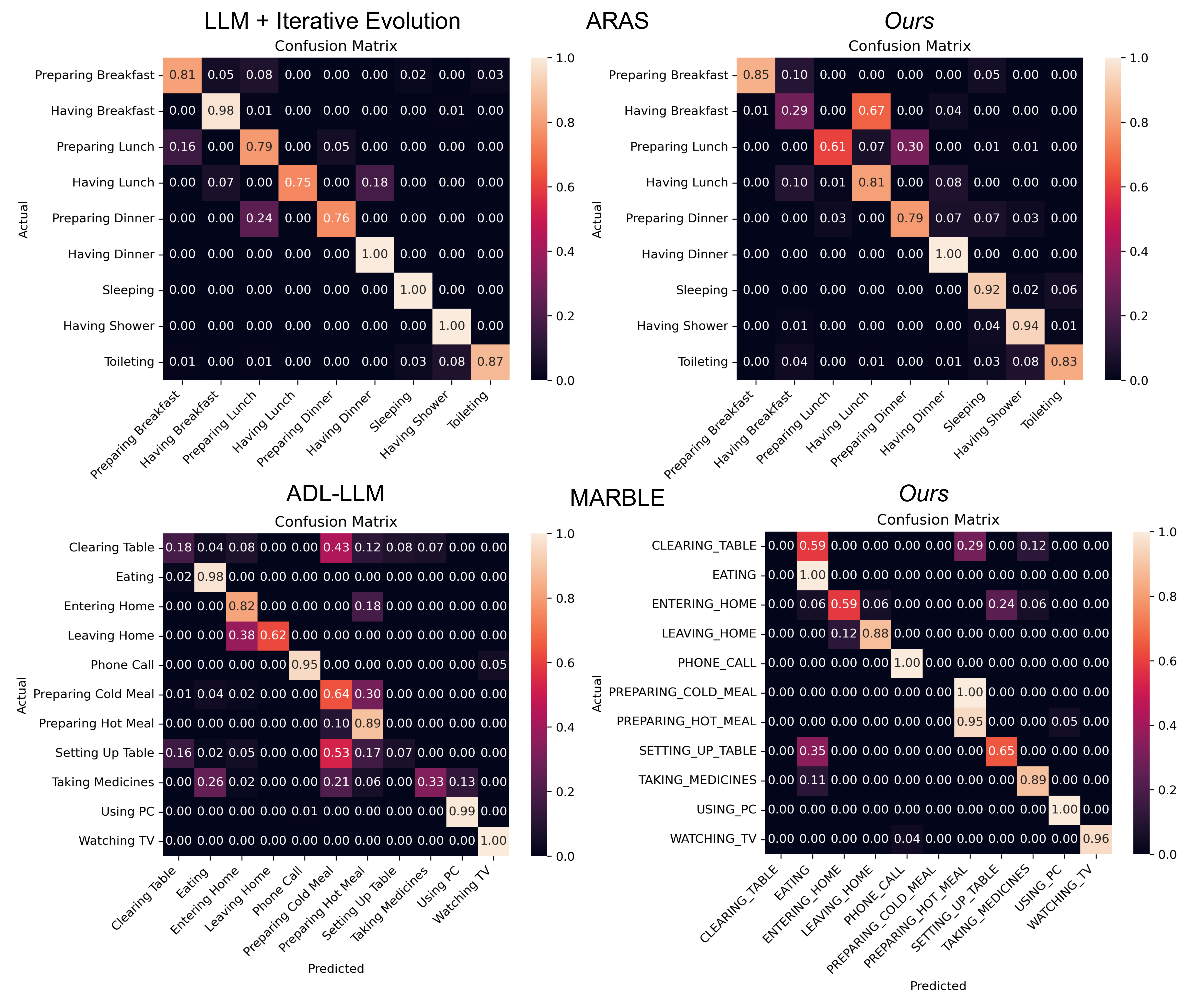}                                            
  \caption{Comparison of our zero-shot approach with two SOTA methods: LLM + Iterative Evolution \cite{gao2024unsupervised} and ADL-LLM \cite{civitarese2024large}. 
  }

\label{fig:aras_marble_comparison}
\end{figure*}

\subsection{Zero-Shot HAR}
\label{subsec:zs_har}
Upon generating datapoint summary $T^X_i$ and activity descriptions $T^Y$, a pre-trained sentence encoder $M$ is used to generate their respective embeddings, where $E^X_i = M(T^X_i)$ and $\{E^Y\} = \{M(T^Y)\}$. 
In this work, we use \textit{all-distilroberta-v1} as our default choice of the sentence encoder for this task, as it is pre-trained on large sentence-level datasets using a self-supervised contrastive objective, making it perform for comparing the similarity between the embeddings of the activity descriptions and the datapoint summary. 
Specifically, we calculate cosine similarities between $E^X_i$ and each $E^Y_j$ and predict the activity label $T^Y_j$ (or essentially $Y_j$), when $j = \arg\max_k \operatorname{sim}(E^X_i, E^Y_k)$. 
This process does not use any labeled/unlabeled data from the smart home, making it zero-shot.


\subsection{Extension to Few-Shot HAR}
\label{subsec:fs_har}
In cases where residents provide a few data samples $\{S_i\}$ and their annotations $\{Y^S_i\}$ to a human-in-the-loop HAR system, the existing zero-shot HAR system must have the ability to be extended to a few-shot solution. 
To address this challenge, we propose a simple idea: we treat the few-shot datapoints as class labels.
In particular, we first generate the summaries $T^S_i$ of the few-shot datapoints as described in Section \ref{subsec:summary_generation} and pass them through the pre-trained sentence encoder $M$ to generate their corresponding embeddings $E^S_i$. 
To perform few-shot HAR, these embeddings are treated as class labels, i.e., for every test datapoint, cosine similarity is calculated between the embeddings of its summary and the set $\{E^Y\} \cup \{E^S_i\}$.
In case a test datapoint window has maximum cosine similarity with $\{E^S_i\}$, the activity label of $\{Y^S_i\}$ gets assigned.

\section{Experiments \& Results}

\begin{figure*}[t]
  \centering
  \includegraphics[width=\textwidth]{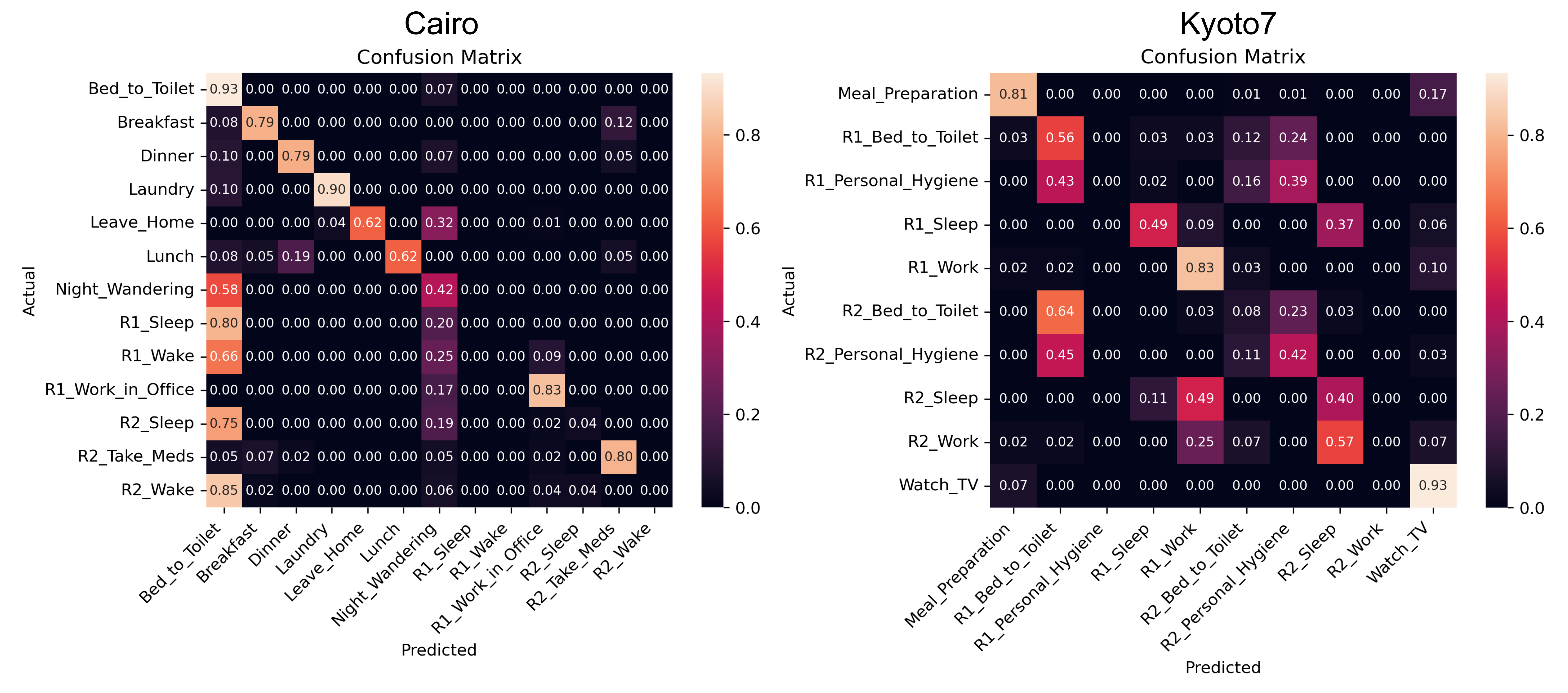}                                            
  \caption{Zero-shot performance on challenging Cairo and Kyoto7 datasets with semantically similar activities of interest.
  }

\label{fig:cairo_kyoto_results}
\end{figure*}

\subsection{Datasets \& Pre-Processing}
We perform our experiments on six publicly available datasets, namely Aruba, Milan, Kyoto7, Cairo, MARBLE, and ARAS \cite{cook2012casas, arrotta2021marble, alemdar2013aras}. 
These datasets contain diverse sensing modalities -- the CASAS datasets (Aruba, Milan, Kyoto7, and Cairo) are collected mainly using event-based motion and door sensors along with a few instances of temperature, switch, and activation readings, whereas, the MARBLE dataset is collected using magnetic, pressure, and switch sensors along with Android application data from smartphones. 
The ARAS dataset is collected using force sensitive
resistors, pressure mats, contact, proximity, sonar distance, temperature sensors, photocells, and infrared (IR) receivers.
The activities of interest collected in these datasets capture aspects of human daily living (eating, sleeping, etc.) as well as healthcare monitoring (bed-to-toilet, taking medicine, etc.).
In addition to these aspects, differences in floor plans, sensor setups, residents (their behaviors and schedules), etc., across the datasets make our evaluation setup comprehensive. 

For the MARBLE and ARAS datasets, we perform the same data pre-processing as mentioned in the SOTA baselines \cite{civitarese2024large, gao2024unsupervised}, which we use for comparison.
Also, we maintain the same testing setup (except for using the ``Other'' class) for the CASAS datasets as mentioned in SOTA prior work \cite{thukral2025layout}.

\subsection{Zero-Shot Performance \& SOTA Comparison}
The performance of our proposed zero-shot approach, in terms of accuracy, weighted and macro F1 scores, is shown in Table \ref{tab:ablations} (first row).
In all the datasets, except for Cairo and Kyoto7, we achieve more than $60\%$ accuracy without using any labeled or unlabeled data for training, which reinforces our approach to be generalizable.
on the MARBLE dataset, we achieve the highest performance with the accuracy nearing $80\%$. 
For the Cairo and Kyoto7 datasets, we refer to the confusion matrices shown in Figure \ref{fig:cairo_kyoto_results}. 
We clearly see that our approach confuses classes that carry similar semantic meaning (e.g., R1\_Sleep, R2\_Sleep, R1\_Wake, R2\_Wake, in Cairo, where the same activity is being performed by different residents). 
Moreover, other confusing activities, such as Bed to Toilet and Personal Hygiene in Kyoto7, contain similar movement patterns, which highlights that our zero-shot approach is capable of correctly classifying sensor data into
semantically similar classes, if not into more granular activities.  

In Figure \ref{fig:aras_marble_comparison}, we compare our approach against two SOTA zero-shot HAR methods, namely LLM + Iterative Evolution \cite{gao2024unsupervised} and ADL-LLM \cite{civitarese2024large}. 
In both cases, we observe comparable performances to the SOTA, without relying on prompting an LLM. 
We also highlight that the misclassifications of our method take place between semantically similar classes, i.e., between Having/Preparing Lunch vs. Dinner, or Preparing Cold/Hot meal.

\subsection{Ablation Experiment}
To ensure that our proposed approach indeed requires all its components and is set at its optimal configuration, we perform an ablation study. 
In this study, we investigate four aspects: the necessity of the (\textit{i}) summary generation module, (\textit{ii}) activity description module, and the optimal choice of (\textit{iii}) sentence encoder, and (\textit{iv}) similarity metric. 
The results from Table \ref{tab:ablations} show a huge performance drop upon removing both the summary generation and activity description modules in all six datasets, building a strong case for including both into our approach. 
Towards choosing the correct sentence encoder, we hypothesize that the one optimized for generating sentence embeddings (all-distilroberta-v1) should be a better choice compared to other use cases. 
Hence, we compare its performance against that of the next closest use case, i.e., paraphrase detection, using paraphrase-distilroberta-base-v2.
We witness a significant drop in performance upon changing the sentence encoder, which supports our original hypothesis. 
Finally, we observe that changing the similarity metric from cosine similarity to L2 norm does not impact the zero-shot performance, which suggests that generating correct embeddings for sensor data and activities is more significant than a more enhanced similarity calculation.

\begin{table*}[t]
\centering
\caption{Ablations experiments on all six datasets by removing the summary generation and activity description components and varying the choice of sentence encoder and similarity metric highlight the optimal design of our proposed solution.}
\begin{adjustbox}{width=\textwidth}
\small
\begin{tabular}{l|ccc|ccc|ccc|ccc|ccc|ccc}
\toprule
\multirow{2}{*}{Method} & \multicolumn{3}{c|}{Aruba} & \multicolumn{3}{c|}{Milan} & \multicolumn{3}{c|}{Cairo} & \multicolumn{3}{c|}{Kyoto7} & \multicolumn{3}{c|}{Marble} & \multicolumn{3}{c}{ARAS} \\
 & Acc. & F1-w & F1-m & Acc. & F1-w & F1-m & Acc. & F1-w & F1-m & Acc. & F1-w & F1-m & Acc. & F1-w & F1-m & Acc. & F1-w & F1-m  \\
\midrule
Ours (proposed) & 0.71  & 0.72 & 0.48 & 0.63 & 0.66 & 0.46 & 0.46 & 0.44 & 0.47 & 0.50 & 0.45 & 0.38 &0.81 & 0.77 & 0.72 & 0.69 & 0.68 & 0.61 \\
w/o \textit{summary} & 0.09  & 0.07  & 0.17  & 0.15 & 0.10 & 0.13 & 0.22  &0.12  & 0.13 & 0.15 & 0.10 & 0.11 & 0.44 & 0.38 & 0.46 & 0.42 & 0.29 & 0.32 \\
w/o \textit{activity descriptors} & 0.02  &0.01  &0.05 & 0.48 & 0.41 & 0.24 &0.24 &0.18 &0.19 &0.16 &0.08 &0.11 &0.36 &0.31 &0.16 &0.36 &0.23 &0.24 \\
w \textit{paraphrase-distilroberta-base-v2} & 0.62 & 0.60 & 0.46 & 0.33 & 0.35 & 0.26 & 0.32 & 0.25 & 0.25 & 0.46 & 0.40 & 0.33 & 0.67 & 0.66 & 0.59 & 0.41 & 0.29 & 0.38 \\
w \textit{L2-Norm} & 0.71 & 0.72 & 0.48 & 0.63 &0.66 &0.46 &0.46 &0.44 &0.47 &0.50 &0.45 &0.38 &0.81 &0.77 &0.72 &0.69 &0.68 &0.61 \\
\bottomrule
\end{tabular}
\label{tab:ablations}
\end{adjustbox}
\end{table*}

\begin{figure*}[t]
  \centering
  \includegraphics[width=\textwidth]{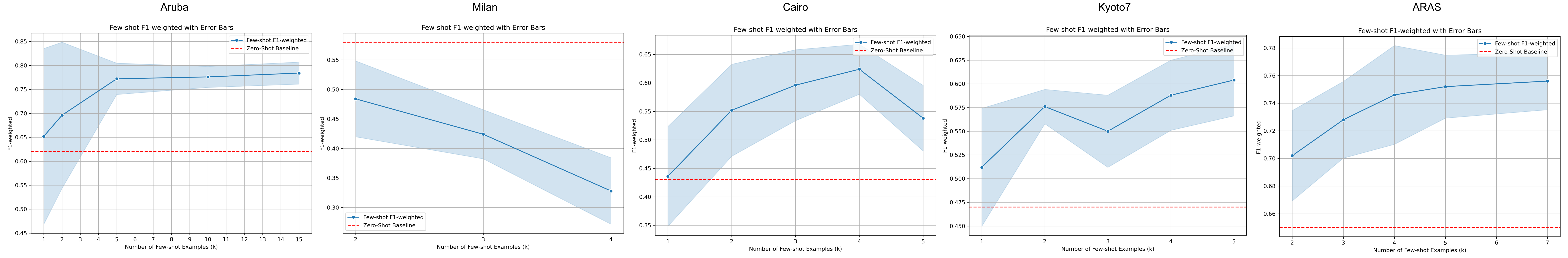}                                            
  \caption{Few-shot performance in terms of the weighted F1 score for varying number of labeled datapoints per class. 
  }

\label{fig:fs_results}
\end{figure*}

\subsection{Few-Shot Performance}
In this experiment, we test the ability of our approach to accommodate a few labeled data samples, if available, towards training and potentially improving the HAR performance.  
In the results shown in Figure \ref{fig:fs_results}, except for the Milan dataset, we observe an overall increase of $\sim13.5\%$ in weighted F1 score compared to the zero-shot baseline (red dotted line), upon using up to 5 labeled datapoints per class. 
However, these results also highlight two aspects of the proposed few-shot recognition method. 
First, in the case of Milan, Cairo, and Kyoto7 datasets, we see cases where increasing the number of labeled samples results in degradation of the performance.
Second, we see a large variance in performance upon randomly selecting the few shots, across 5 random seeds. 
Both these observations point towards the need to develop active learning methods that can \textit{smartly} select the few shots for a guaranteed increase in the performance.

\subsection{Discussion \& Future Directions}
In this work, we proposed a zero-shot HAR method that does not require any training data, labeled or unlabeled, as well as does not require reliance on any outside LLM service, and yet, offers comparable zero-shot performance to the existing SOTA. 
Moreover, our approach is consistent in its structure and applies to a wide range of smart homes, unlike the LLM-based methods that require designing dataset-specific prompts.
Further, we avoid the uncertainty in performance that can be induced by using different LLMs (or their versions) \cite{takeda2024synergistic}. 
Despite this, we observe a few aspects where our method can be potentially improved.

First, the summary generation module in its current state can capture only the fixed and pre-determined aspects of the datapoint -- the challenge of generating datapoint summaries without LLMs remains to be an unsolved challenge. 
Second, our method struggles to classify semantically similar activity classes that may have overlaps in their movement patterns. 
Also, we require very precise definitions of the activity classes for correct recognition -- activities like Wake or Personal Hygiene from Cairo and Kyoto7 datasets cannot be precisely defined, and hence, are poorly classified. 
Finally, our method exhibits the potential to be successfully extended in few-shot scenarios \textit{if} the labeled datapoints for the same are chosen wisely, through an active learning framework. 
Solving these challenges can be an excellent future direction towards making the zero-shot HAR systems work more robustly without using LLMs. 

\section{Conclusion}
Performing HAR in smart homes is a challenging task, where LLMs have proven their efficacy by incorporating commonsense and world knowledge in the recognition pipeline. 
Yet, their promises are limited by concerns regarding privacy, unreliability, and over-reliance. 
To address it, we developed an alternate approach that leverages the embedding space arithmetic of sensor data and activity labels. 
The key lies in generating summaries of datapoints and activity labels at the correct granularity such that embeddings of datapoints lie close to the same as their activity labels. 
Our work provides a comprehensive case study of this approach and builds a case for its further exploration to mitigate the pitfalls of over-reliance on LLMs.

\bibliographystyle{ACM-Reference-Format}
\bibliography{sample-base}


\end{document}